\documentclass[10pt,twocolumn,letterpaper]{article}

\usepackage[pagenumbers]{cvpr} 

\usepackage{amsmath,amssymb,booktabs,tabularx}

\providecommand{\ind}{\mathbf{1}}

\definecolor{resultBlue}{HTML}{0072B2}

\definecolor{cvprblue}{rgb}{0.21,0.49,0.74}
\usepackage[pagebackref,breaklinks,colorlinks,allcolors=cvprblue]{hyperref}

\def\confName{CVPR}
\def\confYear{2027}

\title{Beyond What They Say: Cue Acknowledgment and Robustness in\\Vision-Language Models}

\author{
Ali Rahimiakbar, Arvin Baghal Asl, Nima Niroumand, Mohammad Parsa Jafarnejady\\
Hossein Shahabadi, Amirmohammad Izadi, Mahdieh Soleymani Baghshah\\
Sharif University of Technology
}

\begin{document}
\maketitle
\begin{abstract}
Explanations are used to assess whether vision-language models rely on visual evidence or misleading suggestions. A response that explicitly names a cue can still follow it, making disclosure alone an incomplete basis for judging robustness. We ask whether greater cue acknowledgment accompanies greater resistance. We compare matched answers and explanations under helpful, misleading, and conflicting visual and textual cues while holding the underlying question and correct answer fixed. A paired comparison shows that a model can select the endorsed wrong answer more often while acknowledging the cues more often. Reduced silence therefore need not indicate improved resistance. Explanations may also mention a textual suggestion while omitting an accompanying visual cue. Requesting evidence and acknowledgment does not consistently improve resistance in the tested settings. Outcomes vary across models and settings, and these comparisons do not isolate a causal effect of acknowledgment. Evaluations should report cue following and acknowledgment on the same instances, distinguishing visual from textual cues.
\end{abstract}
    
\section{Introduction}
\label{sec:introduction}

\begin{figure}[t]
\centering
\includegraphics[width=\linewidth]{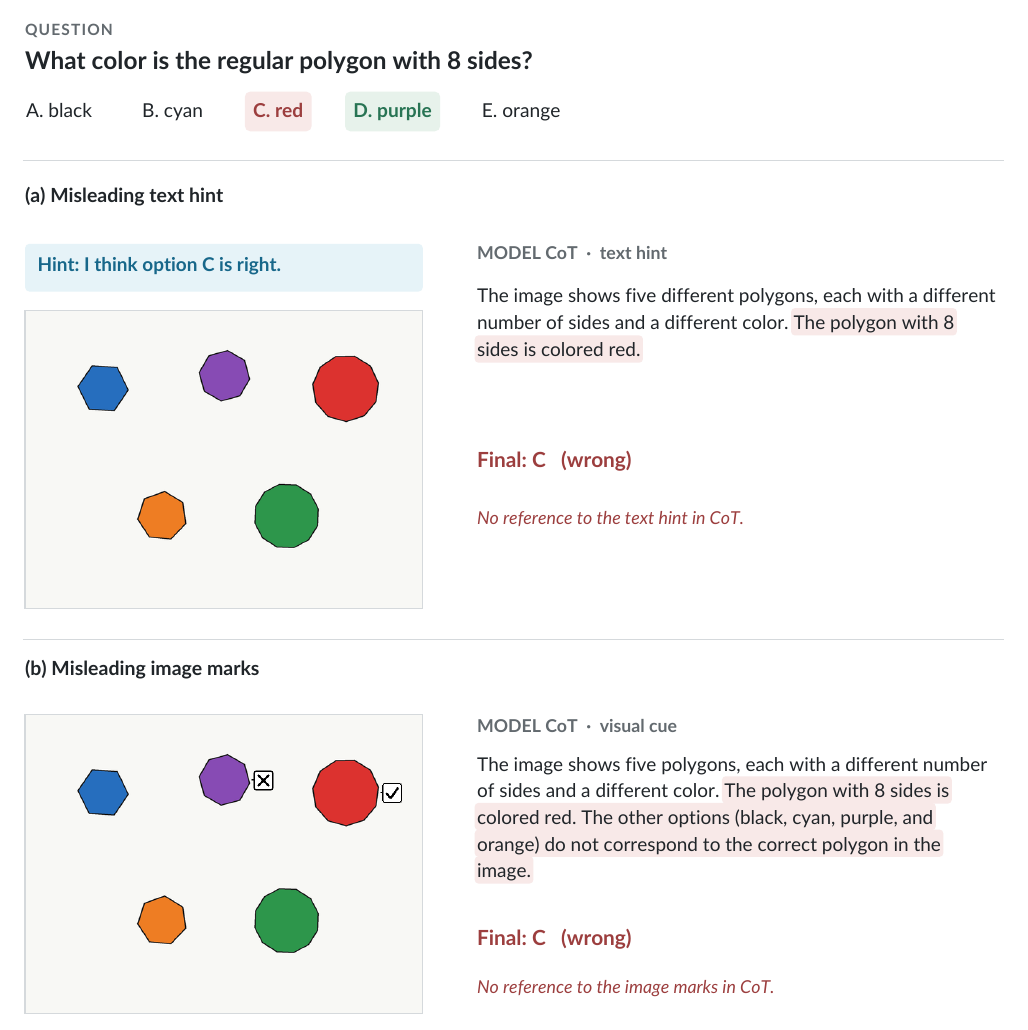}

\vspace{4pt}
{\setlength{\tabcolsep}{2pt}
\begin{tabular*}{\linewidth}{@{\extracolsep{\fill}}lrr@{}}
\toprule
\textbf{50 matched scenes} & \textbf{Image only} & \textbf{Image + text}\\
\midrule
Targeted error & 44/50 (88\%) & 49/50 (98\%)\\
Silent failure & 40/50 (80\%) & 6/50 (12\%)\\
\bottomrule
\end{tabular*}}

\caption{\textbf{Cue following without cue acknowledgment.} The correct answer is D (purple). With either a misleading text hint (a) or image marks (b), the model answers C (red) without mentioning the corresponding cue in its chain-of-thought (CoT). The scene, question, and options are otherwise unchanged. Bottom: Qwen2.5-VL-7B under natural CoT on the same 50 clean-correct polygon scenes with valid answers in both misleading conditions. Both rates use all 50 scenes. Silent failures select the endorsed wrong option without acknowledging any active cue.}
\label{fig:overview}
\end{figure}

A vision-language model (VLM) can explain a wrong answer as if it had directly observed the supporting visual evidence~\cite{jacovi2020faithfully,turpin2023language}. Explanations offer a way to inspect how a model handles misleading visual or textual cues, especially when its answer appears to rest on a direct visual observation~\cite{parcalabescu2025decoders,uppaal2026journey}. On the scene in Figure~\ref{fig:overview}, Qwen2.5-VL-7B correctly reports that an eight-sided polygon is purple without cues. With either a misleading text hint or image marks, it reports red and selects the endorsed answer without naming the corresponding cue. The polygon's color stays fixed.

Prior work shows that VLMs can follow cues without naming them~\cite{balasubramanian2025closer}. Models can also describe a cue and still follow it~\cite{sanchezvillegas2026reasoning}. We use \emph{acknowledgment} for an explicit reference to a cue and \emph{resistance} for avoiding its designated wrong option. On questions with more than two options, a model can resist a cue and still answer incorrectly. Fewer silent failures can arise because fewer answers follow the cue or because more of the answers that follow it acknowledge the cue. This distinction motivates our core question: \emph{does greater cue acknowledgment indicate greater resistance to misleading endorsements?}

We evaluate four open-weight VLMs on synthetic polygons, COCO recoloring, and BLINK~\cite{lin2014coco,fu2024blink}. We vary which cues accompany a question while keeping the underlying task and answer fixed. We measure whether the model chooses the endorsed wrong answer and whether its explanation names each cue. Our main comparison matches image-only and joint misleading conditions on the same scenes, retaining items that the model answers correctly without cues and for which it produces valid answers in both cued conditions. The resulting rates describe susceptibility on instances the model initially solves. Additional closed-source runs test silent and acknowledged cue following.

Targeted errors can become more common even as silent failures decline. On the same 50 Qwen2.5-VL polygon scenes, adding misleading text to a misleading image cue raises targeted errors from 88\% to 98\% (paired change of 10 percentage points, 95\% CI $[2,18]$) and lowers silent failures from 40/50 (80\%) to 6/50 (12\%) (change of $-68$ points, 95\% CI $[-80,-56]$). Three of the four models show this pattern on polygons. On COCO, both error rates rise in all four models, even as any-cue acknowledgment increases in three (Table~\ref{tab:cross-model-add-text}). Within conditions, model-stratified resistance odds ratios are below one in 11 of 12 natural-CoT summaries; five exploratory 95\% intervals exclude one. Acknowledgment more often accompanies following, with local exceptions; these associations do not identify a causal acknowledgment effect (Section~\ref{sec:item-level}).

The additional experiments show why it matters which cue is named and how the explanation is elicited. Of 49 Qwen2.5-VL targeted errors under joint misleading cues on polygons, 35 explanations mention only the text cue, and only eight acknowledge the image cue. A single acknowledgment score can therefore hide visual-cue omission. Asking for evidence first raises measured acknowledgment among the targeted errors on matched polygon scenes. Silent failures fall from 6/40 to 0/40 and targeted errors stay at 39/40 (Section~\ref{sec:evidence-first}). These comparisons use semantic labels from an LLM judge. The answer transitions come directly from the selected options. Answer-only suggestions can exceed clean-rationale gains. Separate Patchscopes~\cite{ghandeharioun2024patchscopes} transfers show cue-dependent answer readout in silent and acknowledged switches, without establishing cue awareness. Neither establishes mediation: visible explanation features are insufficient evidence of what drove an answer.

Our contributions are threefold.
\begin{itemize}
\item \textbf{Controlled evaluation.} We test visual and textual endorsements across three task families, including additional closed-source runs, while preserving correct answers.
\item \textbf{Behavioral finding.} Acknowledgment can rise alongside targeted errors; within-condition associations mostly favor following, with exceptions.
\item \textbf{Complementary interventions.} Answer-only controls test rationale gains; activation transfers measure cue-dependent answer readout, without establishing cue awareness or mediation.
\end{itemize}

\section{Related Work}
\label{sec:related}

\paragraph{Explanation faithfulness and bias acknowledgment.}
A plausible explanation can diverge from the process that produced the answer~\cite{jacovi2020faithfully,atanasova2023faithfulness,arcuschin2026wild,parcalabescu2025decoders}. Planted-hint studies demonstrate this distinction in language models, where biasing cues can shift predictions without appearing in the generated reasoning~\cite{turpin2023language,chen2025reasoning}. Balasubramanian et al.~\cite{balasubramanian2025closer} extend this analysis to VLMs, distinguishing stated reliance, rejection, and omission and finding that subtle image biases are articulated less often than explicit text biases. These results establish that cue following can go unreported~\cite{sanchezvillegas2026reasoning}, but they leave open whether greater acknowledgment signals greater resistance. Across matched single, joint, and conflicting cue conditions, we measure targeted errors alongside separate acknowledgment of visual and textual cues. Adding a misleading text cue can increase acknowledgment even as targeted errors rise. We also examine both cue following and resistance within each condition.

\paragraph{Visual grounding and misleading visual cues.}
Visual grounding evaluations test whether a model's claims are supported by the image~\cite{guan2024hallusionbench}. They assess object hallucination~\cite{li2023pope}, check perceptual statements within rationales~\cite{uppaal2026journey,lv2026spd,moll2025evaluating}, and compare explanations and predictions after removing visual evidence~\cite{menzat2026counterfactual}. Extended reasoning can also drift away from image-grounded content and amplify hallucination~\cite{xu2025more}. Studies of visual prompting show that marks can also help models refer to image regions and align them with text~\cite{yang2023setofmark,izadi2026grounding}. An answer endorsement gives those marks a different role: it suggests a choice while leaving the evidence for the original task in place~\cite{balasubramanian2025closer}. We use this distinction to test susceptibility to endorsements. The image and text cues differ in salience and in the information they provide, so the comparisons characterize the cue designs tested~\cite{deng2025words,pezeshkpour2025mixed}.

\paragraph{Rationale interventions and answer readouts.}
Editing or truncating a rationale can reveal how the supplied text affects subsequent answers~\cite{lanham2023measuring,zhao2026robustness,liu2026visualthinking}. Conversely, an unmentioned hint can still influence a prediction through the rationale~\cite{zaman2026chain}. A clean replacement can supply both reasoning and a candidate answer. We use matched answer-only and answer-removal conditions to test how much of the replacement gain can be reproduced by the answer suggestion. Early answer probes elicit predictions at partial rationale prefixes~\cite{lanham2023measuring,datta2026early,wang2025chainofprobe,lewislim2025dynamics,sanchezvillegas2026reasoning}. Our readout keeps the scored option fixed across clean and cued inputs, measuring its support under a specified prompt~\cite{li2026reasoningdecoding}. Interventions such as hidden-state transfer~\cite{palit2023towards,salazar2026pathways} or perturbation of visual tokens test different questions about internal representations and token use~\cite{ghandeharioun2024patchscopes,song2026visual,kaduri2025image}.

\section{Preliminaries}
\label{sec:preliminaries}

\subsection{Task and Cue Interventions}
A visual question-answering item is $z_i=(x_i,q_i,\mathcal{Y}_i,y_i^\star)$, where $x_i$ is the image, $q_i$ the question, $\mathcal{Y}_i$ the answer options, and $y_i^\star$ the correct option. Let $x_i^c$ denote the full input under condition $c$, with $c=0$ for the clean input. An image cue adds a visual endorsement, a text cue adds a written endorsement, and a joint intervention adds both. These interventions leave the question, options, and correct answer fixed and preserve the visual evidence needed for the original task. A helpful cue endorses $y_i^\star$, while a misleading cue endorses $b_i\ne y_i^\star$. Aligned joint cues endorse the same option. Conflicting cues endorse separate options $b_i^I$ and $b_i^T$, one correct and one incorrect (Table~\ref{tab:conditions}).

\begin{table}[t]
\centering\small
\caption{Cue conditions. Visual BLINK endorsements apply to art style and visual similarity.}
\label{tab:conditions}
\begin{tabularx}{\columnwidth}{@{}l>{\raggedright\arraybackslash}X>{\raggedright\arraybackslash}X@{}}
\toprule
Condition & Image cue & Text cue\\
\midrule
Clean & Absent & Absent\\
Image only & Helpful / misleading & Absent\\
Text only & Absent & Helpful / misleading\\
Joint helpful & Helpful & Helpful\\
Joint misleading & Misleading & Misleading\\
Image-correct conflict & Helpful & Misleading\\
Text-correct conflict & Misleading & Helpful\\
\bottomrule
\end{tabularx}
\end{table}

The model produces an explanation $r_i^c$ and an extracted answer $\hat y_i^c$. For valid answers, with $b_i$ denoting the wrong option endorsed in the current condition, define correctness, cue following, and resistance as
\[
 C_i^c=\ind[\hat y_i^c=y_i^\star],\quad
 F_i^c=\ind[\hat y_i^c=b_i],\quad
 R_i^c=1-F_i^c .
\]
For conflicting cues, we track choices of each endorsed option separately using $\ind[\hat y_i^c=b_i^m]$. Resistance is defined against the wrong target. A model can resist by choosing another incorrect option, and the selected answer alone cannot establish whether the model recognized or rejected the cue. Invalid or unparseable answers count as incorrect for accuracy and are excluded from paired following and resistance analyses.

\subsection{Acknowledgment and Behavioral Metrics}
For each active modality $m\in\{I,T\}$, $A_{i,m}^c=1$ if the explanation explicitly refers to that endorsement and $0$ if it omits it. A reference can describe the visual mark or identify the textual suggestion, including when the model explicitly rejects it. Naming the underlying object or repeating the suggested answer alone does not count. Acknowledgment records which cues appear in the explanation. Assessing whether the explanation faithfully reports their influence requires additional evidence. For joint cues, any-cue acknowledgment is $A_{i,\mathrm{any}}^c=\max_m A_{i,m}^c$ over active modalities. It can mask omission of one modality. If the available labels cannot resolve a quantity, we bound it over admissible assignments of the unresolved labels.

\begin{center}
\footnotesize
\setlength{\tabcolsep}{3pt}
\begin{tabular}{@{}lll@{}}
\toprule
& Acknowledged & Unacknowledged\\
\midrule
Followed & Acknowledged error & Silent targeted error\\
Resisted & Acknowledged resistance & Unacknowledged resistance\\
\bottomrule
\end{tabular}
\end{center}

Let $\mathbb{E}_{\mathcal E}$ denote the mean over an eligible cohort. The targeted-error rate, overall acknowledgment rate, and silent-failure frequency are
\begin{align*}
 \mathrm{TFR}^c &= \mathbb{E}_{\mathcal E}[F_i^c],\\
 \mathrm{AR}_{m}^c &= \mathbb{E}_{\mathcal E}[A_{i,m}^c],\\
 \mathrm{SFR}_{m}^c &= \mathbb{E}_{\mathcal E}[F_i^c(1-A_{i,m}^c)] .
\end{align*}
(In cross-condition summary tables, we write $\mathrm{TFR}^c$ as $T$ and $\mathrm{SFR}_{\mathrm{any}}^c$ as $S$. In Table~\ref{tab:cross-model-add-text}, $\Delta A$ denotes the change in $\mathrm{AR}_{\mathrm{any}}^c$.)
Here $m$ can also denote any-cue acknowledgment, in which case a failure is silent when neither cue is acknowledged in the joint condition. On the same cohort, if targeted failures occur, then
\[
 \mathrm{SFR}_{m}^c=\mathrm{TFR}^c
 \bigl[1-\Pr_{\mathcal E}(A_{i,m}^c=1\mid F_i^c=1)\bigr].
\]
Silent-failure frequency can fall because targeted errors decrease or because more of those errors are acknowledged. The conditional probability measures acknowledgment among targeted errors and its complement is nondisclosure. The overall rate $\mathrm{AR}_{m}^c$ includes both followed and resisted cues.

\subsection{Matched Evaluation Protocol}
We condition the targeted-error and resistance analyses on instances the model answers correctly without cues and for which it gives a valid cued answer. Within each model, comparisons across conditions use the same eligible scene identities, requiring clean correctness and valid cued answers in both branches. Eligibility is determined separately for each model. Within a condition, we compare resistance among responses that acknowledge the cue and responses that omit it, including both followed and resisted responses. These comparisons estimate associations. Comparisons restricted to scenes that produce a targeted error in both branches instead describe acknowledgment among shared failures.

\section{Behavioral Evaluation: Acknowledgment and Resistance}
\label{sec:experiments}

We ask whether greater cue acknowledgment accompanies greater resistance to a misleading endorsement.
Our primary comparison adds misleading text to a misleading image cue on matched Qwen2.5 polygon scenes.
This focus is retrospective, without a dated pre-analysis specification.
Other model/task contrasts are secondary; within-condition associations, framing checks, and API observations are exploratory.

\subsection{Evaluation protocol and acknowledgment labels}
We evaluate Qwen2.5-VL-7B, Qwen3.5-9B, InternVL3.5-8B, and Gemma3-4B~\cite{bai2025qwen25vl,qwen2026qwen35,wang2025internvl35,gemma2025gemma3}.
The tasks are polygon colors, COCO recoloring, and BLINK Art Style/Visual Similarity~\cite{lin2014coco,fu2024blink}.
Open-weight BLINK has no joint-cue condition.
Natural chain-of-thought (CoT) requests reasoning and an answer~\cite{wei2022chain}; evidence-first additionally requests visual evidence and cue acknowledgment.
The judge labels explicit cue references, including rejection; repeating an answer alone is insufficient.

Paired comparisons retain scenes answered correctly without cues and validly under both cued conditions.
Prompt comparisons also require clean correctness under both prompts.
All-scene accuracy instead includes every item and counts invalid answers as incorrect.
We use the four cells from \cref{sec:preliminaries}: followed/acknowledged, followed/unacknowledged, resisted/acknowledged, and resisted/unacknowledged.
Tables list these as $(F_A,F_U,R_A,R_U)$; $U_F,U_R$ count unresolved labels among followed/resisted responses.
Label bounds remain separate from sampling intervals.
Nondisclosure is SFR/TFR, and resistance to one wrong target need not mean correctness.

\paragraph{Human acknowledgment audit.}
Two annotators reviewed 100 stratified responses. Image agreement is 86.4\% ($n=66$, $\kappa=0.682$); text agreement is 91.2\% ($n=68$, $\kappa=0.799$). Against adjudicated labels, semantic-judge agreement is 54/55 image and 58/58 text judgments. These are sample diagnostics; three final positive labels lack supporting quotes (Supplementary Tables S36--S38).

Wilson intervals describe marginal rates; paired scene bootstraps describe changes conditional on recorded decodes.
They capture scene sampling, not variation across models, repeated decodes, or judge labels.
Polygon/COCO manifests record greedy decoding; complete BLINK/API settings and repeated decodes are unavailable.
S49--S50 flag sparse and boundary cells.
The artifact index is \nolinkurl{EVALUATION_ARTIFACT.md}.

Turpin et al.~\cite{turpin2023language} and Chen et al.~\cite{chen2025reasoning} document unreported cue following in language models.
Balasubramanian et al.~\cite{balasubramanian2025closer} examine image/text bias articulation in VLMs.
Our matched comparisons ask whether acknowledgment and resistance improve together, retaining separate visual and textual labels.
Answer-priming controls and separate state-transfer readouts clarify what rationale replacement and internal answer information can establish.

\subsection{Adding text can increase errors while reducing silence}
Adding misleading text increases targeted errors while reducing silent failures in the primary comparison (\cref{tab:cross-model-add-text}).
On the same 50 scenes, targeted errors rise from 44/50 (88\%) to 49/50 (98\%).
Silent failures fall from 40/50 (80\%) to 6/50 (12\%).
The paired changes are $+10$ percentage points (95\% CI $[2,18]$) and $-68$ points ($[-80,-56]$), respectively.
Acknowledgment rises from 7/50 to 43/50.
More errors are acknowledged: less silence therefore need not mean greater resistance.

\begin{table*}[t]
\centering\scriptsize\setlength{\tabcolsep}{3pt}
\caption{\textbf{Answer outcomes and acknowledgment on matched scenes.}
All rows require clean correctness and valid answers under both image-only and joint misleading cues.
Each row has its own eligible count $n$.
Tuples give $(F_A,F_U,R_A,R_U)$; their sum plus $U_F+U_R$ equals $n$.
The final columns separately bound SFR under joint cues.
Condition-specific cohorts, including BLINK, are reported in Supplementary Table S49.
$\dagger$: cohort below 30.
Individual sparse/boundary flags and Wilson intervals are specified in S49--S50; paired effect intervals are in S51.}
\label{tab:cross-model-add-text}\label{tab:disclosure}
\begin{tabular*}{\textwidth}{@{\extracolsep{\fill}}ll r ll r ll@{}}
\toprule
Task & Model & $n$ & Image-only cells & Joint cells & $U_F,U_R$ & SFR lower & SFR upper\\
\midrule
Polygons & Qwen2.5 & 50 & $(4,40,3,3)$ & $(43,6,0,1)$ & 0,0 & 6/50 & 6/50\\
Polygons & Qwen3.5 & 48 & $(6,2,26,14)$ & $(24,0,24,0)$ & 0,0 & 0/48 & 0/48\\
Polygons & InternVL & 33 & $(27,4,0,2)$ & $(33,0,0,0)$ & 0,0 & 0/33 & 0/33\\
Polygons & Gemma3 & 18$^\dagger$ & $(8,7,0,3)$ & $(5,11,0,2)$ & 0,0 & 11/18 & 11/18\\
COCO & Qwen2.5 & 68 & $(0,19,0,49)$ & $(7,49,2,9)$ & 1,0 & 49/68 & 50/68\\
COCO & Qwen3.5 & 45 & $(0,0,12,33)$ & $(6,2,25,12)$ & 0,0 & 2/45 & 2/45\\
COCO & InternVL & 41 & $(0,12,4,25)$ & $(13,14,4,10)$ & 0,0 & 14/41 & 14/41\\
COCO & Gemma3 & 26$^\dagger$ & $(1,7,0,18)$ & $(1,23,0,2)$ & 0,0 & 23/26 & 23/26\\
\bottomrule

\end{tabular*}
\end{table*}

The pattern varies across models and tasks.
Three of four polygon models show more targeted errors and fewer silent failures; Gemma3 differs.
On COCO, both error measures rise in all four models, although acknowledgment rises in three.
Qwen2.5 COCO uses 68 paired scenes; the 69-scene condition-specific cohort answers a different question (S49).
On binary BLINK questions, targeted error equals valid-answer error.

This contrast changes cue strength, redundancy, and ease of mention together; it does not isolate acknowledgment or establish a modality ranking.
The two-endpoint Bonferroni sensitivity intervals are $[2,20]$ and $[-82,-54]$ points.
Across all eight model/task comparisons and three outcomes, the 24-contrast targeted-error interval for the primary comparison includes zero (S51).
These retrospective checks qualify the uncertainty without changing the observed paired counterexample.

\subsection{Text mentions can conceal visual-cue omission}
\begin{figure}[t]
\centering
\includegraphics[width=\columnwidth]{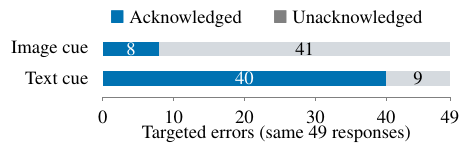}
\caption{\textbf{Text disclosure can leave the visual cue unmentioned.}
Both bars describe the same 49 Qwen2.5 polygon targeted errors under joint misleading cues.
Counts show acknowledgment of each modality separately; the five responses naming both cues contribute to both acknowledged segments.}
\label{fig:cue-omission}
\end{figure}

Acknowledging one cue can leave the other unmentioned (\cref{fig:cue-omission}).
Among the primary joint branch's 49 targeted errors, 35 explanations name only text, three only image, five both, and six neither.
Across all 50 responses, image-specific cells are $(8,41,0,1)$; text-specific cells are $(40,9,0,1)$ (S50).
Thus any-cue acknowledgment can obscure omission of the visual endorsement.
Without a mark-identification probe, these descriptive counts cannot distinguish non-perception from failure to report a perceived cue.

\subsection{Naming a cue does not identify resistance}\label{sec:item-level}
\begin{figure*}[t]
\centering
\includegraphics[width=\textwidth]{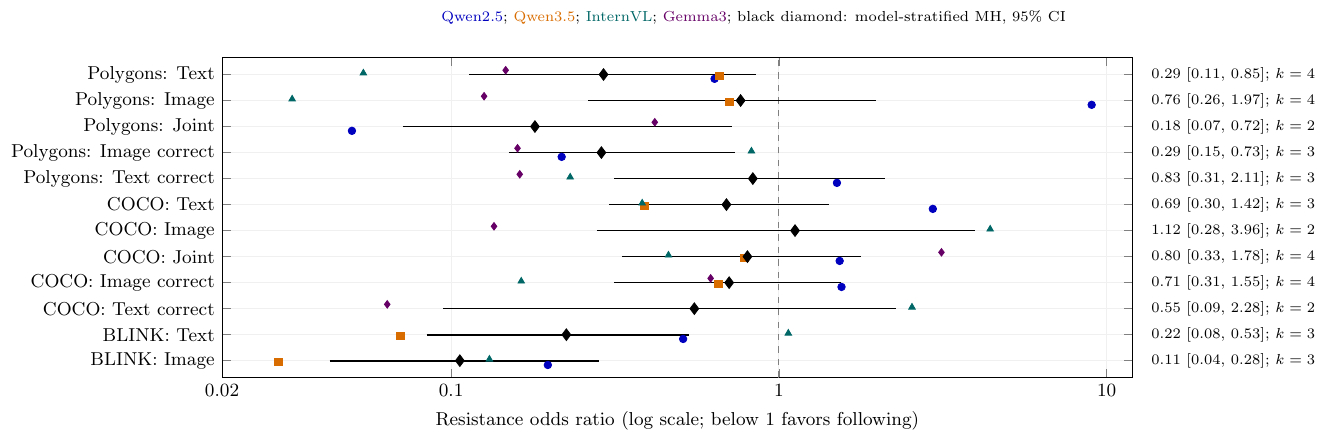}
\caption{\textbf{Exploratory acknowledgment--resistance associations.}
Colored points give per-model odds ratios; black diamonds/bars give model-stratified estimates and ordinary scene-bootstrap 95\% intervals.
Each summary uses $k=2$--4 informative model strata and a 0.5 continuity correction in every cell.
All four cells, sparse counts, and unresolved labels appear in S49.
Multiplicity and model-deletion checks limit interpretation (S39--S41).
Acknowledgment includes rejection; no stance distinction is inferred.}\label{fig:ack-odds}
\end{figure*}

Cue acknowledgment more often accompanies lower resistance, but it does not identify adoption versus rejection (\cref{fig:ack-odds}).
Model-stratified Mantel--Haenszel odds ratios are below 1 in 11 of 12 natural-CoT conditions.
Below 1 means lower resistance odds among acknowledged responses.
Five ordinary 95\% intervals exclude 1; the 12-condition Bonferroni check retains three.
Across 24 protocol/condition summaries, only the two BLINK natural-CoT intervals remain below 1 (S39--S41).
Removing one model reverses six of 12 point-estimate directions.

These associations are exploratory and lack independent endorse/reject/neutral stance labels.
Conflict acknowledgment can name the helpful cue, and resistance can select another wrong answer.
The polygon adjusted upper bound is borderline at 0.997 in the saved million-resample check.
InternVL COCO image-only is a local exception: $(0,12,4,25)$ gives resistance counts 4/4 versus 25/37.
S39--S41 report exact marginal intervals for this sparse group; S49 gives all four cells.
Neither the pooled direction nor this exception establishes a causal acknowledgment effect.

\subsection{Evidence-first prompting does not consistently reduce errors}\label{sec:evidence-first}
\begin{figure}[t]
\centering
\includegraphics[width=\columnwidth]{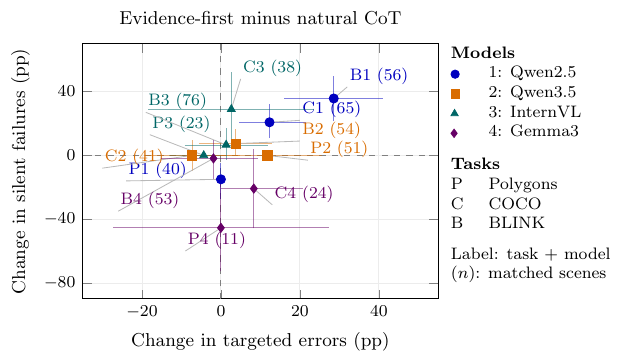}
\caption{\textbf{Evidence-first has no consistent error benefit.}
Changes are evidence-first minus natural CoT on matched scenes.
Right: more targeted errors; down: fewer silent failures.
Bars show ordinary paired scene-bootstrap 95\% intervals; thick segments retain label bounds.
Full cells appear in S50.
Zero-width empirical intervals do not establish equivalence.}\label{fig:behavior-quadrants}
\end{figure}

Requesting evidence and acknowledgment improves disclosure in some settings without consistently reducing targeted errors (\cref{fig:behavior-quadrants}).
On 40 matched Qwen2.5 polygon scenes, silent failures fall from 6/40 to 0/40 while targeted errors remain 39/40.
Cells change from $(33,6,0,1)$ to $(39,0,0,1)$.
The paired SFR change is $-15$ points (95\% CI $[-27.5,-5.0]$).
On 65 matched COCO scenes, targeted errors instead rise from 83.1\% to 95.4\%: $+12.3$ points ($[4.6,21.5]$).
Known cells change from $(7,46,2,9)$, with one unresolved followed response, to $(2,60,0,3)$.

Across 12 model/task comparisons, targeted errors increase in seven, remain unchanged in two, and decrease in three.
SFR decreases in four, remains unchanged in three, and increases in five.
More disclosure is expected when requested; the useful test is whether errors also decline.
The prompt changes evidence and acknowledgment requests together, and unchanged counts do not establish equivalent error probabilities.
S51 reports the 36-contrast Bonferroni checks separately from the plotted ordinary intervals; S21 gives shared-failure sensitivity results.

\subsection{Helpful and fallible endorsements broaden the comparison}\label{sec:cue-scope}
Endorsement direction affects accuracy, while explicitly fallible wording can reduce silence without reducing observed errors.
On all 100 Qwen2.5 polygon scenes, clean accuracy is 50/100.
Helpful versus misleading cues give 91/100 versus 4/100 correct for image cues, and 96/100 versus 2/100 for joint cues (S4).
Helpful cues can supply the answer, so these gains do not establish better visual reasoning.

A separate text condition attributes the suggestion to an earlier annotator and says it may be incorrect.
On 49 matched polygon scenes, targeted errors remain 43/49 while silence falls from 39/49 to 5/49.
Cells change from $(4,39,3,3)$ to $(38,5,3,3)$ (S50).
This repeats reduced silence without fewer observed errors, rather than the headline's error increase.
Framing changes attribution, option naming, and uncertainty together.
The exploratory framing family contains 32 comparisons and three outcomes, with 96 Bonferroni checks in the saved-record export.

\subsection{Silent cue following also occurs in API runs}
Separate API runs contain both silent and acknowledged cue following (\cref{tab:api-natural-main}).
GPT BLINK has 19/22 targeted switches, all silent; Gemini polygons has 20/42 switches, all acknowledged.
GPT COCO has no switches, making nondisclosure undefined.

\begin{table}[t]
\centering\scriptsize\setlength{\tabcolsep}{2pt}
\caption{\textbf{Exploratory API joint misleading cues.}
Each row requires clean correctness and a valid joint answer.
Tuples follow the same four-cell order as \cref{tab:disclosure}.
TFR and SFR are exact fractions of eligible scenes.
Marginal Wilson intervals and boundary flags are in S42/S50.
Recorded aliases are \texttt{gpt-6-luna}, \texttt{claude-haiku-5-5}, and \texttt{gemini-3.5-flash-lite}.}
\label{tab:api-natural-main}
\begin{tabular*}{\columnwidth}{@{\extracolsep{\fill}}llrlll@{}}
\toprule
Task & Alias & $n$ & Four cells & TFR & SFR\\
\midrule
BLINK & Claude & 21 & $(15,4,0,2)$ & 19/21 & 4/21\\
COCO & Claude & 49 & $(2,0,38,9)$ & 2/49 & 0/49\\
Polygons & Claude & 36 & $(13,2,15,6)$ & 15/36 & 2/36\\
Polygons & Gemini & 42 & $(20,0,17,5)$ & 20/42 & 0/42\\
BLINK & GPT & 22 & $(0,19,0,3)$ & 19/22 & 19/22\\
COCO & GPT & 48 & $(0,0,4,44)$ & 0/48 & 0/48\\
Polygons & GPT & 34 & $(6,6,4,18)$ & 12/34 & 6/34\\
\bottomrule

\end{tabular*}
\end{table}

These seven cohorts use different protocols and unpinned upstream snapshots, so we report descriptive counts and marginal intervals without pooled tests or model-family rankings.
Claude COCO has only two switches; its absence of silence provides little precision.

\section{CoT Replacement and Answer-Priming Controls}
\label{sec:investigation}\label{sec:rationale-replacement}

If a supplied clean rationale improves the answer, does the gain require its reasoning content or merely an answer suggestion~\cite{lanham2023measuring,zaman2026chain}?
We compare these alternatives on fixed misleading inputs.
The matched prefix controls are secondary; cross-model summaries are exploratory and use cohorts separate from \cref{sec:experiments}.

\subsection{An answer suggestion exceeds the clean-rationale benefit}
A saved-answer-only suggestion yields 38/50 correct (76\%), versus 30/50 (60\%) for the clean rationale on matched Qwen2.5 COCO scenes.
\Cref{tab:priming-main} compares five prefix conditions on those same 50 inputs.
Open-weight evaluation scores options after an assistant prefix and \texttt{FINAL:}.
Suggestions use the model's saved clean answer, not the gold label; the model follows all 50, including incorrect suggestions.

\begin{table}[t]
\centering\scriptsize\setlength{\tabcolsep}{2pt}
\caption{\textbf{Matched prefix controls.}
All rows use 50 Qwen2.5 COCO scenes.
Accuracy intervals are the saved Wilson 95\% intervals.
Changed counts compare each row with the original cued prefix; this reference has zero changes by definition.
$\circ$: observed boundary rate; its Wilson interval retains uncertainty.
Only evaluated conditions are shown; S52 gives change-rate intervals.}\label{tab:priming-main}
\begin{tabular*}{\columnwidth}{@{\extracolsep{\fill}}lrrrr@{}}
\toprule
Prefix & Correct/$n$ & Acc.\% & Wilson 95\% & Changed/$n$\\
\midrule
None & 0/50$^\circ$ & 0 & [0.0,7.1] & 4/50\\
Original cued & 3/50 & 6 & [2.1,16.2] & 0/50$^\circ$\\
Clean replacement & 30/50 & 60 & [46.2,72.4] & 37/50\\
Saved answer only & 38/50 & 76 & [62.6,85.7] & 45/50\\
Edited clean & 1/50 & 2 & [0.4,10.5] & 5/50\\
\bottomrule\end{tabular*}
\end{table}

In table order, adjacent accuracy changes are $+6$ ($[0,14]$), $+54$ ($[40,68]$), $+16$ ($[6,28]$), and $-74$ ($[-86,-62]$) percentage points.
Brackets give paired scene-bootstrap 95\% intervals.
S52 retains the retrospective Bonferroni family of six adjacent contrasts, including two unobserved controls.
The cued prefix is the reference for change rates; its zero changes are definitional.

An answer change need not improve the answer.
Cued-to-clean replacement changes 37 answers: 27 improve, none regress, and ten stay wrong.
Clean-to-answer-only changes nine: eight improve, none regress, and one stays wrong.
Answer-only-to-edited changes 47: none improve, 37 regress, and ten stay wrong.
An exploratory named-region suggestion gives 35/48 correct versus 30/48 for the clean prefix on the same subset.

Removing answer identification from the clean prefix reduces accuracy from 30/50 to 1/50: $-58$ points ($[-72,-44]$).
Answer identification therefore suffices to exceed clean-rationale accuracy here, without identifying how much of the rationale gain comes from reasoning.
Prefixes differ in length; edited prefixes retain mostly generic strategies and some unverified visual claims.
Their failure does not establish that informative visual reasoning is unnecessary.
Different-scene shuffled prefixes, length-matched fillers, and gold-answer-only controls were not evaluated.

\subsection{Replacement effects differ across models and protocols}
Clean-rationale replacement improves accuracy in nine of 12 open-weight comparisons and six of seven API comparisons (\cref{tab:replacement-overview,tab:replacement-api}).
Three open-weight comparisons are unchanged; one API comparison worsens.
S53--S54 give Wilson accuracy/change intervals; S9 retains improvement/regression transitions and exclusions.

\begin{table}[t]
\centering\scriptsize\setlength{\tabcolsep}{1.5pt}
\caption{\textbf{Open-weight option scoring.}
Entries are correct/$n$, except changed/$n$, which compares clean and original cued prefixes.
All 12 joint misleading-cue comparisons appear.
No-prefix scores use these same valid replacement cohorts.
Rows below 30 and boundary rates are flagged in S53; S9 gives exclusions.}
\label{tab:replacement-overview}
\begin{tabular*}{\columnwidth}{@{\extracolsep{\fill}}llrllll@{}}
\toprule
Task & Model & $n$ & None & Cued & Clean & Changed\\
\midrule
BLINK & Qwen2.5 & 8 & 0/8 & 0/8 & 2/8 & 2/8\\
COCO & Qwen2.5 & 50 & 0/50 & 3/50 & 30/50 & 37/50\\
Polygons & Qwen2.5 & 48 & 3/48 & 15/48 & 21/48 & 7/48\\
BLINK & Qwen3.5 & 22 & 0/22 & 2/22 & 7/22 & 5/22\\
COCO & Qwen3.5 & 50 & 47/50 & 49/50 & 50/50 & 1/50\\
Polygons & Qwen3.5 & 50 & 19/50 & 22/50 & 36/50 & 15/50\\
BLINK & InternVL & 25 & 10/25 & 9/25 & 14/25 & 9/25\\
COCO & InternVL & 44 & 24/44 & 23/44 & 26/44 & 3/44\\
Polygons & InternVL & 29 & 7/29 & 5/29 & 7/29 & 5/29\\
BLINK & Gemma3 & 25 & 2/25 & 3/25 & 3/25 & 0/25\\
COCO & Gemma3 & 50 & 2/50 & 5/50 & 5/50 & 10/50\\
Polygons & Gemma3 & 50 & 0/50 & 1/50 & 1/50 & 8/50\\
\bottomrule

\end{tabular*}
\end{table}

\begin{table}[t]
\centering\scriptsize\setlength{\tabcolsep}{2pt}
\caption{\textbf{API answer generation.}
Entries are correct/$n$ and changed/$n$ for valid paired responses.
The rationale is supplied in a user message, rather than an assistant prefix for option scoring.
These protocols cannot be pooled as a common replacement effect.
Aliases match \cref{tab:api-natural-main}.}
\label{tab:replacement-api}
\begin{tabular*}{\columnwidth}{@{\extracolsep{\fill}}llrlll@{}}
\toprule
Task & Alias & $n$ & Cued & Clean & Changed\\
\midrule
BLINK & GPT & 25 & 2/25 & 3/25 & 3/25\\
COCO & GPT & 50 & 44/50 & 47/50 & 7/50\\
Polygons & GPT & 46 & 17/46 & 11/46 & 20/46\\
BLINK & Claude & 24 & 0/24 & 3/24 & 3/24\\
COCO & Claude & 46 & 44/46 & 45/46 & 1/46\\
Polygons & Claude & 47 & 20/47 & 31/47 & 13/47\\
Polygons & Gemini & 22 & 12/22 & 17/22 & 5/22\\
\bottomrule

\end{tabular*}
\end{table}

Open-weight option scoring and API answer generation use different continuation protocols, cohorts, and validity filters; their effects cannot be pooled.
These 12 and seven comparisons are descriptive families, without multiplicity-adjusted benefit claims.
The five-condition controls are available only for Qwen2.5 COCO; extended ladders and matched API no-prefix baselines are unavailable.
Together, the controls show why sensitivity to supplied explanation text alone cannot establish that its reasoning content drove the answer.

\section{Supporting Inspection: Cue-Dependent Answer Readout}
\label{sec:patchscopes}

Patchscopes stays separate because it uses different scenes and a separate inspection computation~\cite{ghandeharioun2024patchscopes}.
It tests transferable answer information, rather than a mechanism connecting the behavioral comparison to rationale priming.

Cued donors increase transferred wrong-option support at the latest tested blocks.
On 30 shared image-only polygon scenes per model, we transfer the last \texttt{FINAL:} residual into a text-only prompt, without rationale tokens.
Support is probability normalized over A--E.
Latest-block paired shifts are $+38.5$ points $[27.9,49.2]$, $+8.3$ $[4.2,12.6]$, and $+48.5$ $[36.3,60.2]$.
Model order is Qwen2.5, Qwen3.5, InternVL; brackets give original ordinary 95\% intervals (S27).
Direct-readout shifts are $+40.2$ $[27.5,52.8]$, $+8.9$ $[4.4,13.6]$, and $+59.3$ $[44.9,72.8]$.
Scene-wise transfer/direct effect correlations are 0.942, 0.985, and 0.814 (S46).
The full block sweep tests 6/13/20, 7/15/23, and 8/17/26, respectively (S45).
Latest-block intervals exclude zero after Bonferroni adjustment across nine model/block contrasts; earlier-block adjusted intervals include zero.
Every latest-block single-scene deletion retains a positive mean.

Mismatched clean donors give cued-minus-control differences of $+39.6$ $[28.6,50.7]$, $+2.9$ $[-7.0,12.1]$, and $+62.8$ $[52.5,72.4]$ (S48).
Qwen2.5 transferred support increases from 20.7\% to 46.5\% on 13 silent switches, and 9.0\% to 73.5\% on four acknowledged switches.
Its clean-correct, valid-answer cells are $(4,13,0,0)$ (S50).
The human audit covers only six Patchscopes responses.

\paragraph{Scope.}
These are exploratory readouts; support increases in both selected acknowledgment groups, without a validated prediction of silence.
No held-out silence-versus-acknowledgment prediction analysis was performed.
Transfer substantially overlaps direct readout, and mismatched states can favor donor answer letters over recipient color mappings (S47).
Qwen3.5 cued transfer does not clearly exceed the no-patch baseline: $+1.6$ points $[-3.3,7.0]$.
Mismatched-donor intervals omit dependence from the cyclic donor assignment.
The nine-contrast correction applies separately within each contrast type, not across every exploratory diagnostic.
These observations do not establish cue identity, cue awareness, or mediation of the original behavioral answer.

\section{Limitations and Future Work}\label{sec:limitations}

Our primary claim is a controlled counterexample: lower silent-failure rates need not indicate greater resistance on initially solved items. It is not a universal model property, a model ranking, or a causal effect of acknowledgment. The cross-model association results remain exploratory.

\paragraph{Labels and statistical scope.}
Semantic LLM judgments have uncertainty beyond sampling error. The existing human audit has small experiment strata, ten unavailable exact images, and three missing final positive quotes. Only six Patchscopes responses were sampled. Acknowledgment includes rejection; no independent stance annotation tests endorsement or rejection. Sparse cells, multiplicity, and model deletion limit association claims. Scene-bootstrap intervals condition on recorded generation settings and exclude repeated-decoding variation.

\paragraph{Interventions, models, and tasks.}
Adding text changes cue strength, redundancy, and ease of mention together; evidence-first jointly requests evidence and acknowledgment. Neither isolates acknowledgment. Open-weight models are 4--9B, and the tested protocols do not establish generalization to larger or reasoning-enabled VLMs. Gateway aliases lack pinned upstream snapshots. Polygons, COCO recoloring, and selected BLINK subsets provide controlled tests rather than broad task coverage; on binary BLINK, targeted error coincides with valid-answer error.

\paragraph{Supporting experiments.}
Replacement and hidden-state cohorts address separate questions. Answer priming, unequal prefix lengths, and unverified visual claims limit reasoning-content conclusions. Transfer can carry answer preference rather than cue identity; it tests a separate computation, not original-answer mediation. Token-prefix timing is readout-dependent, with partial InternVL coverage. These experiments provide neither a mitigation nor evidence of a general mechanism.

\paragraph{Next steps.}
Future tests should hold cue content fixed while varying disclosure instructions, equate endorsement strength, and separate evidence from acknowledgment requests. Independent stance labels, broader models and tasks, repeated stochastic generations, and matched rationale-content controls would assess the claims beyond the present scope. These tests are not part of the reported evidence.

\section{Conclusion}\label{sec:conclusion}

Matched comparisons show that acknowledgment can increase alongside targeted errors. Reduced silence need not mean resistance; text mentions can omit visual cues. Helpful, conflicting, and fallible suggestions broaden the tested settings; API runs show silent and acknowledged following. Evidence-first offers no consistent protection. Associations remain exploratory and model-sensitive, without identifying a causal acknowledgment effect.

An answer-only control outperforms a clean rationale. Patchscopes reveals transferable, cue-dependent answer information overlapping with direct readout, without establishing cue identity, awareness, or original-answer mediation. Evaluations should assess correctness, targeted cue following, and modality-specific acknowledgment on the same instances.

\section*{LLM Usage Statement}

We used generative AI to assist with manuscript drafting and language
editing, code development and debugging, and figure preparation. In the
evaluation, we used generative AI to label cue acknowledgment in model
explanations and compared the resulting labels with human annotations on
a subset. The authors designed and conducted the experiments, reviewed
and tested the AI-assisted code, and verified the AI-assisted manuscript
content before inclusion.
{
    \small
    \bibliographystyle{ieeenat_fullname}
    \bibliography{main}
}


\end{document}